\pdfoutput=1

\documentclass[11pt]{article}

\usepackage[preprint]{acl}

\usepackage{times}
\usepackage{latexsym}

\usepackage[T1]{fontenc}

\usepackage[utf8]{inputenc}

\usepackage{microtype}

\usepackage{inconsolata}

\usepackage{graphicx}

\usepackage{amsmath}
\usepackage{amssymb}
\usepackage{xcolor}
\usepackage{multirow} 
\usepackage{booktabs}
\usepackage{multicol}
\usepackage{colortbl}
\usepackage{tikz,pgfplots}
\usepackage{tcolorbox}
\usepackage{algorithm}
\usepackage{algpseudocode}
\usepackage{hyperref}
\usepackage{silence}
\pgfplotsset{compat=1.18}
\title{I0T: Embedding Standardization Method Towards Zero Modality Gap}

\author{Na Min An\thanks{\space{ } Equal contribution}~~ Eunki Kim\footnotemark[1]~~ James Thorne\thanks{\space{ } Corresponding author}~~ Hyunjung Shim\footnotemark[2]
\\
 KAIST AI
\\
 \{naminan, eunkikim, thorne, kateshim\}@kaist.ac.kr
}

\begin{document}
\maketitle

\begin{abstract}
Contrastive Language-Image Pretraining (CLIP) enables zero-shot inference in downstream tasks such as image-text retrieval and classification. However, recent works extending CLIP suffer from the issue of \textit{modality gap}, which arises when the image and text embeddings are projected to disparate manifolds, deviating from the intended objective of image-text contrastive learning. We discover that this phenomenon is linked to the modality-specific characteristic that each image/text encoder independently possesses and propose two methods to address the modality gap: (1) a post-hoc embedding standardization method, $\text{I0T}_{\text{post}}$ that reduces the modality gap approximately to zero and (2) a trainable method, $\text{I0T}_{\text{async}}$, to alleviate the modality gap problem by adding two normalization layers for each encoder. Our I0T framework can significantly reduce the modality gap while preserving the original embedding representations of trained models with their locked parameters. In practice, $\text{I0T}_{\text{post}}$ can serve as an alternative explainable automatic evaluation metric of widely used CLIPScore (CLIP-S). The code is available in \url{https://github.com/xfactlab/I0T}.
\end{abstract}

\section{Introduction}

Utilizing Vision-language models (VLMs) such as Contrastive Language-Image Pretraining (CLIP) \cite{radford2021learning} has been a common practice for performing multimodal tasks \cite{goel2022cyclip, furst2022cloob, li2023scaling, zhang2024long, gao2024softclip, sarto2023positive, hu2023infometic, lee2024fleur}. Despite these successes, CLIP and its variants \cite{xu2021videoclip, zhang2022contrastive, goel2022cyclip, sarto2023positive, zhang2024long} suffer from a significant limitation known as the \textit{modality gap}; image and text embeddings diverge in the latent space, projected to separate manifolds \cite{liang2022mind, fahim2024its}. This is in contrast to the original image-text contrastive learning (CL) objective, which pulls and pushes the positive and negative pair of image and text embeddings \cite{radford2021learning}, deviating from the shared statistical model representing reality \cite{huh2024position}.

\begin{figure}[t!]
    \centering
    \includegraphics[width=\columnwidth]{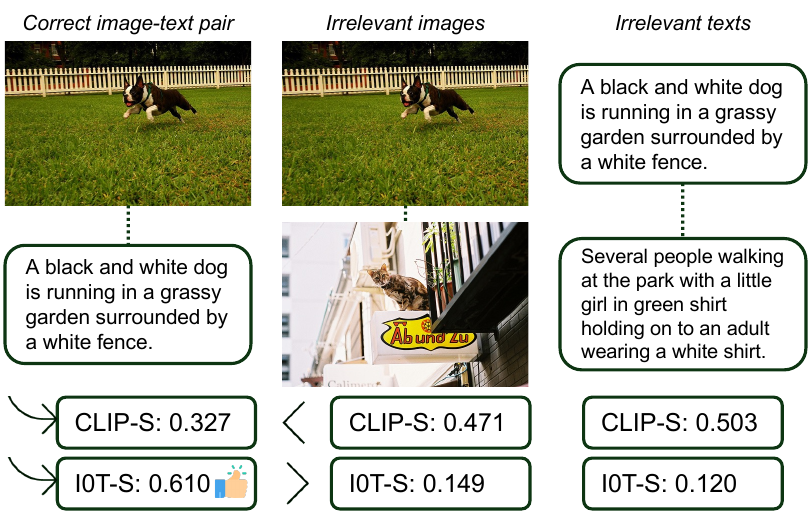}
    \caption{Refined scoring system using our proposal (I0T-S) than CLIP-S. I0T-S assigns a higher similarity score for the correct image-text pair than irrelevant pairs.}
    \label{fig:clipscore_i0tscore}
\end{figure}

The undesirable symptom of \textit{modality gap} is that data within the same modality always have higher semantic similarity than the cross-modal data. Therefore, CLIP cannot draw an accurate semantic relationship for the data pool mixed with different modalities. This problem is especially noticeable when CLIP is extended as an automatic evaluation metric, widely used CLIPScore (CLIP-S) \cite{hessel2021clipscore, sarto2023positive}, which measures the cosine similarity between image and text embeddings. Figure~\ref{fig:clipscore_i0tscore} shows that CLIP-S unintuitively returns a lower score for the correct image-text pair than the irrelevant image-image and text-text pairs due to embedding discrepancy between images and texts.

Prior approaches to mitigate the modality gap have focused on shifting \cite{liang2022mind} or training \cite{fahim2024its, eslami2024mitigate} the embeddings of the positive pairs closer together. However, they did not attempt to find and attribute explicit factors in the image and text embeddings that lead to the modality gap. In contrast, we find the actual attributing factor of the modality gap; CLIP inadvertently learns the inherent characteristic of each modality (referred to as \textit{modality-specific characteristic} in this paper), inducing similar activation patterns within the normalized embeddings of all different images (or texts) from each image (or text) encoder. These patterns, characterized by peak activations with distinct negative and positive directions for image and text embeddings (later visualized in Figure~\ref{fig:post_hoc_methods}), significantly contribute to the modality gap. We find that it is crucial to discard not only these peak activations on a specific few dimensions but also existing modality-specific characteristics across all dimensions from each encoder to mitigate the modality gap.

Here we propose a framework, Zero (\textbf{0}) Modality Gap between \textbf{I}mage-\textbf{T}ext embedding representations (\textbf{I0T}) that aims to minimize the modality gap towards zero. Correspondingly, it is also crucial to maintain rich semantic embedding representations, even if they become closely aligned. The first stage of I0T is a plug-in-play module that can be implemented with any readily available fine-tuning strategies. The second stage of I0T can be addressed with two proposed approaches. We first develop $\text{I0T}_{\text{post}}$ that standardizes the normalized image and text embedding activations independently by subtracting the mean vectors of each modality and renormalizing with Frobenius normalization on the frozen encoders from the first stage. 

$\text{I0T}_{\text{post}}$ offers a more explainable image captioning evaluation metric than CLIPScore \cite{hessel2021clipscore} (referred to as I0TScore (I0T-S) in Figure~\ref{fig:clipscore_i0tscore}) by assigning a similar range of scores for across different modalities and within the same modality, attributable to the low modality gap property. However, this post-hoc embedding standardization method needs a sufficient amount of data samples with a similar distribution as a test set; hence, we present  $\text{I0T}_{\text{async}}$ that learns the aligned embeddings with no access to the test distribution. Our main contributions can be summarized as follows:

\begin{itemize}
\setlength\itemsep{-0.1em}
\item Achieving both modality gap and downstream performances is challenging; yet, we propose an I0T framework that significantly reduces the gap without hurting performances.

\item $\text{I0T}_{\text{post}}$ and $\text{I0T}_{\text{async}}$ significantly reduce the modality gap while enhancing text-to-image retrieval scores by 9.2\% and 6.7\%.

\item We are the first to propose an automatic evaluation metric, I0TScore, that can be applied to data across different modalities, overcoming the limitation of CLIPScore that only works within a single modality.
\end{itemize}

\begin{figure*}[t!]
    \centering
    \includegraphics[width=\textwidth]{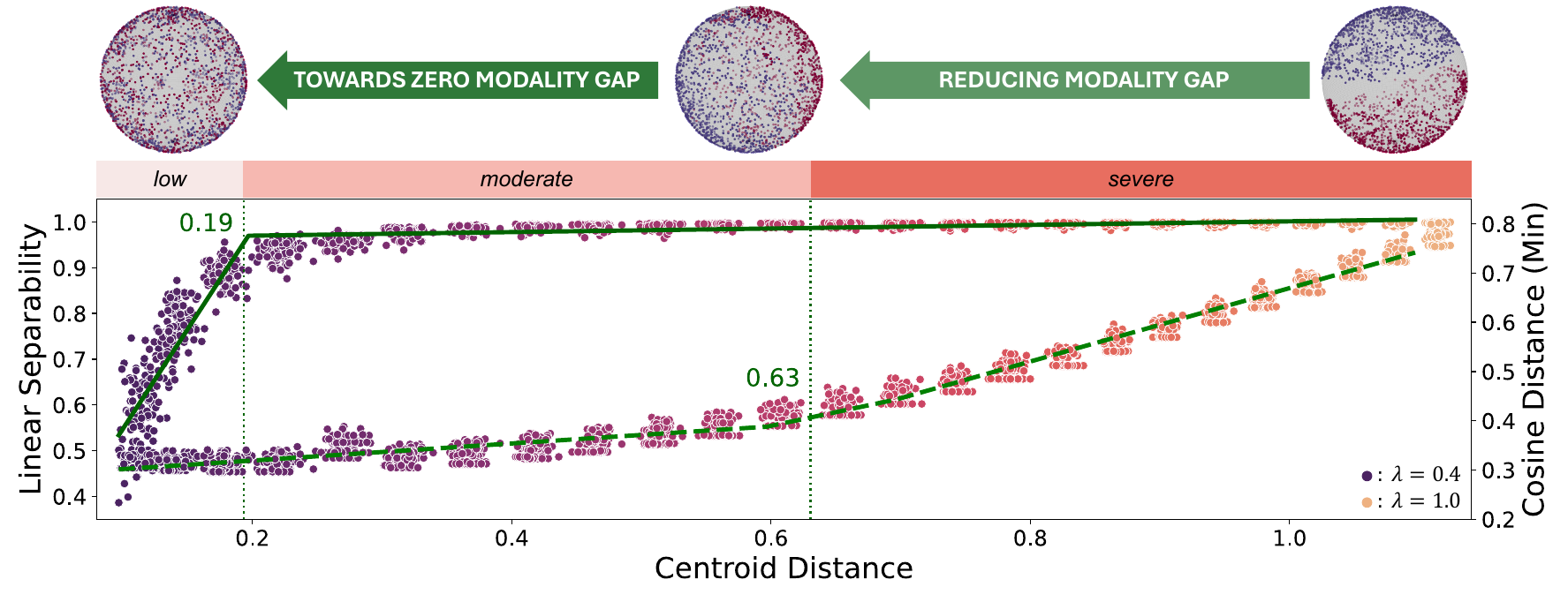}
    \caption{Linear separability and minimum cosine distance (dashed line) \textit{vs}. centroid distance illustrated with corresponding 3D-projected embeddings. The embeddings are categorized by three modality gap severity levels: \textit{severe}, \textit{moderate}, and \textit{low}.}
    \label{fig:severity_level}
\end{figure*}

\section{Related Works}

\subsection{CLIP-Based Models}
Vision-language models (VLMs) have addressed multimodal tasks that require a joint understanding of visual and textual data \cite{liu2024visual, li2022blip}. Most modern VLMs utilize CLIP-style architectures due to CLIP's exceptional performance in zero-shot downstream tasks using pre-trained image and text encoders \cite{radford2021learning, jia2021scaling}. However, CLIP alone shows limitations in producing consistent representations \cite{goel2022cyclip}; Hence, CyCLIP \cite{goel2022cyclip} reduces the similarity of mismatched pairs of image and text (cross-modal cyclic) and the image pairs and the corresponding text pairs (in-modal cyclic). Long-CLIP \cite{zhang2024long} uses knowledge-preserved enlarged positional embedding, handling up to 248 input tokens, significantly greater than the 77 tokens restricted in CLIP. FLIP \cite{li2023scaling} proposes a technique where a significant portion of image patches is randomly masked during training. SoftCLIP \cite{gao2024softclip} uses softened target labels derived from fine-grained intra-modal self-similarity.

\subsection{Modality Gap}

The issue of the modality gap is pervasive in VLMs such as CLIP, caused by embeddings for images and texts occupying disjoint regions in the latent space \cite{liang2022mind, oh2024geodesic}. This gap, by definition, restricts the model from utilizing the entire latent space. The root cause of this modality gap has been debated: \citealp{ramasinghe2024accept} claims that the intrinsic differences between image and textual data unavoidably result in the modality gap. \citealp{liang2022mind} attributes the gap to the resulting narrow cone due to the high model hidden dimension. \citealp{fahim2024its} suggests that the gap, mainly caused by the contrastive learning objective, could be reduced with additional loss terms for uniformity and stricter cross-modal alignment \cite{wang2020understanding}. In this work, we are interested in \textit{removing} the actual attributing factor of the gap, in contrast to \textit{accepting} the modality gap \cite{ramasinghe2024accept} to extend CLIP as an explainable evaluation metric.

\section{Preliminary Analyses}

\textit{Modality gap} was introduced by \citealp{liang2022mind} and is defined as the centroid distance (CD) between the mean of normalized image embeddings ($\mathbf{x}_{i} \in \mathbb{R}^{d}$, $i = 1, 2, ..., n$) and mean of normalized text embeddings ($\mathbf{y}_{i} \in \mathbb{R}^{d}$, $i = 1, 2, ..., n$). Formally, $\triangle_{\text{CD}} := {||\bar{\textbf{x}}-\bar{\textbf{y}}||}_{F}$, where $\bar{\textbf{x}} := \frac{1}{n}\sum_{i=1}^{n} \mathbf{x}_{i}$, $\bar{\textbf{y}} := \frac{1}{n}\sum_{i=1}^{n} \mathbf{y}_{i}$, with $d$ and $n$ representing the model's hidden dimension and the data size. \citealp{fahim2024its} quantify the gap as the linear separability (LS) of image and text embeddings \cite{shi2023towards}. To measure LS, or $\triangle_{\text{LS}}$, we divide the given dataset into training (70\%) and test (30\%). Then, we train a linear regression model and report 1 $-$ mean squared error of the model separability of image and text embeddings, following the same procedure as \citealp{fahim2024its}.

\begin{figure*}[t!]
    \centering
    \includegraphics[width=\textwidth]{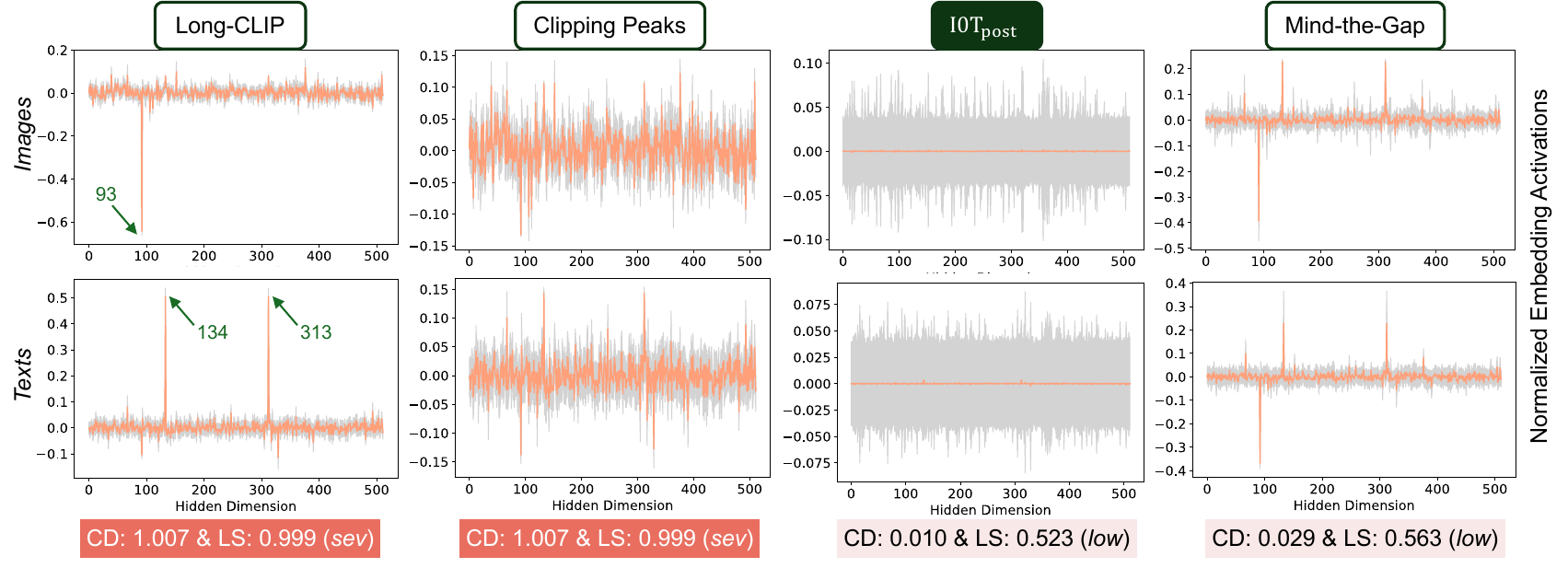}
    \caption{Comparison of normalized embedding activations (avg: salmon, std: gray) and modality gap across three post-hoc methods applied on Long-CLIP.}
    \label{fig:post_hoc_methods}
\end{figure*}

\subsection{Severity Levels of Modality Gap}
To integrate the different definitions of the modality gap, we characterize the relationship between CD, LS, and minimum cosine distance\footnote{We subtract maximum cosine similarity corresponding to top-1 predicted labels from 1.0 for all samples in the dataset.} (MCD; $\triangle_{\text{MCD}}$) using piece-wise linear interpolation (Figure~\ref{fig:severity_level}). We find that if $\triangle_{\text{CD}}$ $<$ 0.19, $\triangle_{\text{LS}}$ deviates from 1.0. Also, as $\triangle_{\text{CD}}$ $>$ 0.63, MCD increases with a steeper slope than the slope in $\triangle_{\text{CD}}$ $<$ 0.63 (\textit{See} Appendix~\ref{app:categ} for details). Thus, our categorization of the modality gap using a relationship of CD, LS, and MCD is as follows:

\begin{itemize}
\setlength\itemsep{-0.1em}
\item \textit{\textbf{Severe}}: $\triangle_{\text{CD}}$ $\geq$ 0.63
\item \textit{\textbf{Moderate}}: 0.19 $\leq$ $\triangle_{\text{CD}}$ $<$ 0.63
\item \textit{\textbf{Low}}: $\triangle_{\text{CD}}$ $<$ 0.19
\end{itemize}

\subsection{Normalized Embedding Activations}
The attributing factor of the modality gap observed in CLIP can be informed through our analysis of the normalized embedding activations\footnote{These embeddings with Frobenius norms set to 1 are used for calculating the cosine similarity of an input image and text.} from each image/text encoder. We first investigate distinct peak activations in the normalized image and text embeddings and then theoretically show that these peak activations contribute to the modality gap. As displayed in the first column of Figure~\ref{fig:post_hoc_methods} (or Figure~\ref{fig:outlier_activations} in Appendix~\ref{app:norm_act}), a similar pattern of normalized embedding activations is shown across the hidden dimensions for different images and texts with a small standard deviation. Also, we consistently observe negative peak activations at the 93rd dimension for \textit{all} image samples and positive peak activations at the 134th and 313th dimensions for \textit{all} text samples with low standard deviation, regardless of semantic representations of each sample per modality. This phenomenon is possibly due to one of the root causes of the modality gap discussed in Related Works. This suggests that each encoder encodes modality-specific characteristics that can contribute to the embedding discrepancy between images and texts. Thus, mitigating these modality-specific characteristics across all dimensions, particularly peak activations at a few dimensions, is essential to alleviate the modality gap.

\subsection{Contribution to Modality Gap}
We now demonstrate how these peak activations in the normalized image and text embeddings prevent the cosine similarity from reaching high values. To illustrate the upper bound of the cosine similarity, suppose there exists one negative peak, $p$, in normalized image activation ($\textbf{x}_i = [x_1, x_2, ..., x_d]$) and two positive peaks of $q$ in normalized text embedding activations ($\textbf{y}_i = [y_1, y_2, ..., y_d]$), and $|p| \gg x_i$ and $|q| \gg y_i$, in align with our empirical finding (Figure~\ref{fig:post_hoc_methods}). For simplicity, we assume that the other non-peak activations are uniformly distributed. Then, the upper bound of $|\text{cos}(\textbf{x}_i, \textbf{y}_i)|$ converges to $\sqrt{(1-p^2)(1-2q^2)}$ as $d \rightarrow \infty$ (proof in the Appendix~\ref{app:proof}). If we set $p$ to be $-\frac{1}{2}$, and $q$ to be $\frac{1}{3}$ (Long-CLIP activations from Figure~\ref{fig:post_hoc_methods}), the upper bound of $|\text{cos}(\textbf{x}_i, \textbf{y}_i)|$ converges to 0.76. Since this converged value is less than 1, it implies that the existence of peak activations hinders the cosine similarity of text and image embeddings from being close to 1, inducing a modality gap.

\section{Methodology}

The I0T framework consists of two stages, the initial stage being a plug-in-play module that can be skipped if the user only wants to tackle the modality gap problem of the models. The second stage of I0T is applied asynchronously \textit{after} the first stage. The motivation behind these divided stages is maintaining the semantic representations by locking the model parameters in the first stage and mitigating the modality gap in the following stage.

\subsection{The First Stage of I0T}

In this initial stage, our goal is to enhance the semantic representations of CLIP. Since our work is the first to present a 2-step paradigm that handles both the semantic representations and the modality gap problem, we share our best strategies to improve overall downstream performances on CLIP using a mixture of recently introduced CLIP fine-tuning strategies from several works of literature. We follow the implementation of \textbf{Long-CLIP} \cite{zhang2024long}, but with a key difference; we find that using only long captions for alignment (\textbf{Long-CLIP-only}) on \textbf{COCO} from the ShareGPT4V dataset \cite{chen2023sharegpt4v} significantly reduces the training time ($\sim$1/10) while achieving better performances in downstream tasks (refer to Appendix~\ref{app:exp_deta} for details).

We also use a combination of the standard contrastive learning and \textbf{Cyclic} losses \cite{goel2022cyclip}, $\mathcal{L}_{\text{CyCLIP}} := \mathcal{L}_{\text{CLIP}} + 
0.25 \mathcal{L}_{\text{I-Cyclic}} + 0.25 \mathcal{L}_{\text{C-Cyclic}}$. We fine-tune CLIP for three epochs using the AdamW optimizer \cite{loshchilov2018decoupled} with a learning rate 1e-6 and a weight decay 1e-2. We set a batch size of 128 (64 for each GPU device), and we use the standard contrastive learning loss with the temperature log scale of 4.6052. We apply this procedure to all comparison methods to ensure a comparison procedure. All the details of comparison baselines and evaluation downstream tasks are in Appendix~\ref{app:exp_deta}.

\subsection{The Second Stage of I0T}

\paragraph{Post-hoc Method to Reduce Modality Gap} To mitigate the modality gap, it is crucial to remove modality-specific characteristics from the embeddings of each encoder. An intuitive approach might involve reducing peak activations through \textit{clipping}. However, we find that clipping the normalized activations within the range of $-$0.1 to 0.1 and re-normalized with Frobenius norms still results in \textit{severe} modality gap (\textit{see} the second column in Figure~\ref{fig:post_hoc_methods}). We hypothesize that the unclipped activations might still encompass the property that commonly exists across the activations for each image/text encoder linked to the modality gap.

Motivated by the limitation of this clipping method, we develop an embedding standardization method to remove modality-specific characteristics from the normalized activations across entire dimensions. We standardize the normalized embedding activations ($\textbf{x}_i, \textbf{y}_i \in \mathbb{R}^{d}$) by subtracting the mean vectors ($\bar{\textbf{x}}, \bar{\textbf{y}} \in \mathbb{R}^{d}$) for each modality and re-normalize them by dividing by the Frobenius norms (the third column of Figure~\ref{fig:post_hoc_methods}): $\textbf{x}_{i}^{'} = \text{Normalize}(\textbf{x}_{i} - \bar{\textbf{x}}), \textbf{y}_{i}^{'} = \text{Normalize}(\textbf{y}_{i} - \bar{\textbf{y}})$

We can clearly observe that our post-hoc method significantly reduces the modality gap, similar to the post-hoc shifting method of Mind-the-Gap (MG) \cite{liang2022mind} (compare the third and the fourth columns of Figure~\ref{fig:post_hoc_methods}); however, with no outlier peaks and mean activations close to zero across all hidden dimensions, suggesting that our approach can remove the actual cause factor of the modality gap including, especially peak activations, unlike the MG approach (compare peak activations between third and fourth columns in Figure~\ref{fig:post_hoc_methods}).

\paragraph{Learnable Method to Reduce the Gap} Although $\text{I0T}_{\text{post}}$ significantly reduces the modality gap, it does not support zero-shot inference for a single sample. To overcome this, we explore a method to \textit{automatically} reduce the modality gap without relying on post-hoc refinement. The key point of our $\text{I0T}_{\text{async}}$ method is to add an independent batch normalization (BN) layers, $\text{BN}_{\text{img}}$ and $\text{BN}_{\text{txt}}$ for each encoder. This enables the model to learn the means and variances of normalized image and text embedding activations without affecting the semantic encoding capability of the encoder (\textit{see} Discussion). Through this process, the model iteratively updates the running means and variances of normalized embedding activations for each modality: $ \bar{\textbf{x}}_{t+1} = \alpha \bar{\textbf{x}}_{\mathcal{B}} + (1 - \alpha) \bar{\textbf{x}}_{t}, \bar{\textbf{y}}_{t+1} = \alpha \bar{\textbf{y}}_{\mathcal{B}} + (1 - \alpha) \bar{\textbf{y}}_{t}$.

$\bar{\textbf{x}}_{t+1}$ and $\bar{\textbf{y}}_{t+1}$ denote the updated running means in training time step $t+1$, incorporating the batch mean vectors, $\bar{\textbf{x}}_{\mathcal{B}} = \sum_{i=1}^{m}{\textbf{x}_{i}}$ and $\bar{\textbf{y}}_{\mathcal{B}} = \sum_{i=1}^{m}{\textbf{y}_{i}}$, with averaging factor $\alpha = 0.1$, and batch size, $m=64,128,256,512$. We use the final updated running mean of normalized image and text embedding activations, $\bar{\textbf{x}}_{\text{train}} := \bar{\textbf{x}}_{T}$, $\bar{\textbf{y}}_{\text{train}} := \bar{\textbf{y}}_{T}$ ($T$: final training step) as the \textit{learned} modality-specific characteristics of images and texts. Similarly, the final updated running variance of normalized image and text embedding activations are $\sigma_{\textbf{x}_\text{train}} := \sigma_{\textbf{x}_\text{T}}$, $\sigma_{\textbf{y}_\text{train}} := \sigma_{\textbf{y}_\text{T}}$, which are empirically observed as close to 1.0 across all $d$ dimensions.  The final updated image and text semantic representations can be expressed as ($\epsilon$ = 1e-05): $\textbf{x}_{i}^{'} = \text{Normalize}(\textsl{\textbf{W}}_{\text{img}}(\frac{\textbf{x}_{i} - \bar{\textbf{x}}_{\text{train}}}{\sqrt{\sigma_{\textbf{x}_\text{train}} + \epsilon}}) + \textsl{\textbf{b}}_{\text{img}})$ and $ \textbf{y}_{i}^{'} = \text{Normalize}(\textsl{\textbf{W}}_{\text{txt}}(\frac{\textbf{y}_{i} - \bar{\textbf{y}}_{\text{train}}}{\sqrt{\sigma_{\textbf{y}_\text{train}}  + \epsilon}}) + \textsl{\textbf{b}}_{\text{txt}}) $, where $\textsl{\textbf{W}}_{\text{img}}, \textsl{\textbf{W}}_{\text{txt}} \in \mathbb{R}^{d}$ and $\textsl{\textbf{b}}_{\text{img}}, \textsl{\textbf{b}}_{\text{txt}} \in \mathbb{R}^{d}$ indicate the weights and biases of $\text{BN}_{\text{img}}$ and $\text{BN}_{\text{txt}}$.

It is crucial to consider how we can effectively train the weights/biases of these BN layers of $\text{I0T}_{\text{async}}$. We follow the exact training implementation details as the first stage with $\mathcal{L}_{\text{CyCLIP}}$ as the loss objective (AddBatchNorm $=$ False in Appendix~\ref{app:alg} Algorithm~\ref{alg:cap}). Then, we freeze the weights of the fine-tuned encoders to preserve semantic representations and train the BN layers asynchronously afterward (AddBatchNorm $=$ True in Appendix~\ref{app:alg} Algorithm~\ref{alg:cap}).

When training these BN layers, here, we propose Multimodal Contrastive Learning of Sentence and Image Embeddings (MCSIE), our re-implementation of MCSE \cite{zhang2022mcse}, a learning method using an unsupervised-positive augmentation. Unlike MCSE, dropout (rate: 0.1) is applied to every multi-head attention layer of both the image encoder (ViT-B/32, \citealp{dosovitskiy2020image}) and the text encoder (Transformer, \citealp{vaswani2017attention}), augmenting relations between all combinations of images/augmented images and texts/augmented texts with $\sum_{\mathcal{E}^{I} \in \{\mathcal{I}, \mathcal{I}_{\text{aug}}\}, \mathcal{E}^{T} \in \{\mathcal{T}, \mathcal{T}_{\text{aug}}\}} \mathcal{L}$, where $\mathcal{L}$ indicates a loss function. From our ablation study (see Table~\ref{tab:mcsie} in Appendix~\ref{app:add_exp}), we find MCSIE effectively further reduces the modality gap, suggesting that it enables BNs to learn the modality-specific characteristics robustly.

\begin{table*}[t!]
\begin{center}
\resizebox{0.91\textwidth}{!}{%
\begin{tabular}{c|c|ccc|cc|cc|cc}
\toprule
\multicolumn{1}{c|}{} & \multicolumn{1}{c|}{} & \multicolumn{3}{c|}{Modality Gap} & \multicolumn{6}{c}{Downstream Performances} \\
 
\multicolumn{1}{c|}{Models} & \multicolumn{1}{c|}{$\#$} & Centroid & Linear & Sev. & \multicolumn{2}{c|}{Retrieval $\uparrow$} & \multicolumn{2}{c|}{Classification $\uparrow$} & \multicolumn{2}{c}{Relative Cor. $\uparrow$} \\

\multicolumn{1}{c|}{} & \multicolumn{1}{c|}{Par $\downarrow$} & Dist. $\downarrow$ & Sep. $\downarrow$ & Level $\downarrow$ & \multicolumn{1}{c}{I2T} & \multicolumn{1}{c|}{T2I} & \multicolumn{1}{c}{CIFAR} & \multicolumn{1}{c|}{Bird} & \multicolumn{1}{c}{Expert} & \multicolumn{1}{c}{CF} \\
 
\midrule
Long-CLIP & 353m & 0.9904 & 0.9998 & \cellcolor[HTML]{EC7063}\textit{sev} & 71.90 & 75.00 & \textbf{65.03} & \underline{5.43} & 53.21 & 34.86 \\ 
LCO & 353m & 0.9965 & 0.9997 & \cellcolor[HTML]{EC7063}\textit{sev} & 72.50 & 75.90 & 64.46 & \textbf{5.81} & 51.42 & 35.17 \\ 
LCCO & 353m & 1.0070 & 0.9999 & \cellcolor[HTML]{EC7063}\textit{sev} & \textbf{74.90} & \underline{76.10} & \underline{64.75} & 4.90 & \textbf{54.57} & 35.43 \\ 
LCCOM & 353m & 0.9682 & 0.9998 & \cellcolor[HTML]{EC7063} \textit{sev} & 73.70 & 74.60 & 64.17 & 4.93 & 54.55 & \textbf{35.72} \\ \midrule
+ LN & 354m & 1.0068 & 0.9999 & \cellcolor[HTML]{EC7063} \textit{sev} & \underline{74.70} & \textbf{76.20} & 64.58 & 5.08 & \underline{54.56} & 35.42 \\ 
+ $\text{LN}^{*}$ & 353m & 0.9696 & 0.9998 & \cellcolor[HTML]{EC7063} \textit{sev} & 73.60 & 74.10 & 64.36 & 4.87 & 54.52 & \underline{35.70} \\ 
+ BN & 354m & \underline{0.5285} & \underline{0.9983} & \cellcolor[HTML]{F5B7B1} \textit{mod} & 71.60 & 70.10 & 63.14 & \textbf{5.81} & 53.36 & 32.53 \\ 
\midrule
+ $\text{BN}^{*}$ (I0T$_{\text{async}}$) & 354m & \textbf{0.4795} & \textbf{0.9960} & \cellcolor[HTML]{F5B7B1} \textit{mod} & 72.50 & 73.80 & 62.97 & 5.21 & 53.33 & 33.08 \\
\bottomrule
\end{tabular}}
\caption{Comparison of modality gap and downstream tasks performances across variations of Long-CLIP-based models. $^{*}$ indicates the asynchronous training method where we fine-tuned the frozen encoders from the first stage. The \textbf{bolded} and \underline{underlined} values indicate the best and the second-best performances.}
\label{tab:ablation_comparison}
\end{center}
\end{table*}
\begin{table*}[!t]
\begin{center}
\resizebox{0.95\textwidth}{!}{%
\begin{tabular}{c|c|ccc|cc|cc|cc|c}
\toprule
\multicolumn{1}{c|}{} & \multicolumn{1}{c|}{} & \multicolumn{3}{c|}{Modality Gap} & \multicolumn{6}{c|}{Downstream Performances} & \multicolumn{1}{c}{} \\
 
\multicolumn{1}{c|}{Models} & \multicolumn{1}{c|}{$\#$} & Centroid & Linear & Sev. & \multicolumn{2}{c|}{Retrieval $\uparrow$} & \multicolumn{2}{c|}{Classification $\uparrow$} & \multicolumn{2}{c|}{Relative Corr. $\uparrow$} & \multicolumn{1}{c}{Rank $\downarrow$} \\

\multicolumn{1}{c|}{} & \multicolumn{1}{c|}{Par $\downarrow$} & Dist. $\downarrow$ & Sep. $\downarrow$ & Level $\downarrow$ & \multicolumn{1}{c}{I2T} & \multicolumn{1}{c|}{T2I} & \multicolumn{1}{c}{CIFAR} & \multicolumn{1}{c|}{Bird} & \multicolumn{1}{c}{Expert} & \multicolumn{1}{c|}{CF} & \multicolumn{1}{c}{}\\

\midrule
 
CLIP & 353m & 0.7642 & 0.9985 & \cellcolor[HTML]{EC7063} \textit{sev} & 69.60 & 67.10 & \textbf{65.05} & \textbf{5.94} & 51.00 & \underline{34.30} & 3.88 \\
$\text{MG}_{\lambda=0.375}$ & 353m & \underline{0.0291} & \underline{0.5632} & \cellcolor[HTML]{FDEDEC} \textit{low} & 45.40 & 54.40 & 43.26 & 1.67 & 42.84 & 29.26 & 5.63 \\
$\text{MG}_{\lambda=0.5}$ & 353m & 0.2493 & 0.9858 & \cellcolor[HTML]{F5B7B1} \textit{mod} & 38.10 & 46.50 & 44.13 & 1.37 & 39.70 & 27.25 & 6.63 \\
$\text{MG}_{\lambda=-0.5}$ & 353m & 1.3799 & 0.9998 & \cellcolor[HTML]{EC7063} \textit{sev} & 45.20 & 54.40 & 9.52 & \underline{5.35} & NaN & NaN & 7.50 \\
CLOOB & 354m & 0.4832 & 0.9899 & \cellcolor[HTML]{F5B7B1} \textit{mod} & 69.60 & 72.60 & 60.40  & 4.91 & 50.06 & 31.71 & 4.50 \\
Unif-Align & 353m & 0.4636 & 0.9921 & \cellcolor[HTML]{F5B7B1} \textit{mod} & 51.20 & 46.00 & 50.32 & 3.83 & 42.47 & 29.31 & 5.88 \\
PAC-S & 353m & 0.7583 & 0.9990 & \cellcolor[HTML]{EC7063} \textit{sev} & \underline{72.60} & 71.60 & 58.61 & 3.74 & \textbf{54.00} & \textbf{36.10} & 4.50 \\ 
\midrule
I0T$_{\text{async}}$ & 354m & 0.4795 & 0.9960 & \cellcolor[HTML]{F5B7B1} \textit{mod} & 72.50 & \underline{73.80} & 62.97 & 5.21 & 53.33 & 33.08 & \underline{3.63} \\ 
I0T$_{\text{post}}$ & 353m & \textbf{0.0102} & \textbf{0.5374} & \cellcolor[HTML]{FDEDEC} \textit{low} & \textbf{73.30} & \textbf{76.30} & \underline{63.07} & 4.76 & \underline{53.97} & 33.58 & \textbf{1.88} \\
\bottomrule
\end{tabular}}
\caption{Comparison of modality gap and downstream performances across different ViT-B/32-based CLIP models. The \textbf{bolded} and \underline{underlined} values indicate the best and the second-best performances.}
\label{tab:main_comparison}
\end{center}
\end{table*}

\section{Results}

We first present how we build our best CLIP semantic representations implementing the first stage of I0T on conventionally conducted tasks - image-text retrieval and image classification. Unlike previous works, we also test how CLIP-based models correlate with human ranking scores. Then, we evaluate the effectiveness of our final I0Ts (\textit{i.e.}, I0T$_{\text{async}}$ and I0T$_{\text{post}}$) on two perspectives: (1) modality gap and (2) downstream performances since it is crucial to maintain semantic representations even if the embeddings are aligned. Most importantly, we show the applicability of I0T$_{\text{post}}$ as an automatic reference-free evaluation metric.

\subsection{Reducing Modality Gap while Maintaining Semantic Representaions}
In Table~\ref{tab:ablation_comparison}, we show how we develop the final I0T$_{\text{async}}$ starting from Long-CLIP \cite{zhang2024long}, but again, note that this can be easily replaced with other new CLIP fine-tuning strategies to be introduced as the future work. Long-CLIP-only (LCO) significantly reduces the training time by using only COCO from all training datasets used in Long-CLIP with similar overall downstream performances. Long-CyCLIP-only (LCCO) adds only the cyclic losses \cite{goel2022cyclip} to LCO but scores higher overall downstream performances. Long-CyCLIP-only + MCSIE (LCCOM), however, shows a slight decrease in downstream performances, but it also shows a slight decrease in modality gap, showcasing the possibility of reducing the gap using MCSIE; however, all these methods still show \textit{severe} modality gap.

Adding separate layer normalization for each encoder either during the fine-tuning (+LN) or after training (+LN*) still results in a severe modality gap. In contrast, our proposed way of adding independent batch normalization layers for each encoder either during the fine-tuning (+BN) or after training (+BN*) results in a \textit{moderate} level of modality gap. We find that asynchronous training strategy (+BN*) performs better than non-asynchronous training strategy (+BN) in terms of both modality gap and retrieval performances; thus, proposing the BN* as our final I0T$_{\text{async}}$. Similarly, our final $\text{I0T}_{\text{post}}$ is designed based on LCCOM, incorporating its strengths for optimal performance.

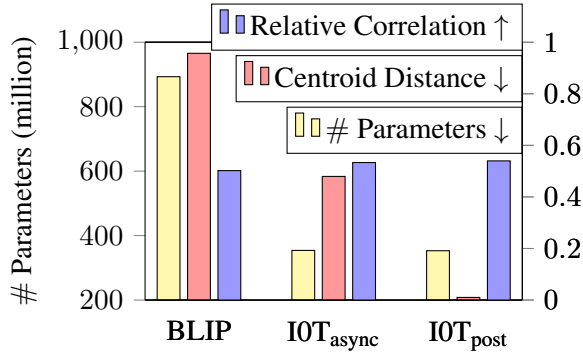
\begin {figure}[t!]
\centering
\begin{tikzpicture}
\begin{axis}[
    width  = 0.85\columnwidth,
    height = 5cm,   
    ybar,
    bar width = 3.0mm,
    bar shift = -0.4cm,
    enlarge x limits = 0.20,
    ylabel style = {align=center},
    ylabel = $\#$ Parameters (million),
    ymin = 200,
    ymax = 1000,
    ytick = {200, 400, 600, 800, 1000},
    ytick pos = left,
    ytick align = outside,
    symbolic x coords = {BLIP,I0T$_{\text{async}}$,$\text{I0T}_{\text{post}}$},
    axis y line*=left,
    xtick = data,
    xtick style={draw=none},
    xtick pos = bottom,
    xtick align= outside,
    xticklabel style = {rotate=0, text width=2.0cm, align=center},
    xlabel style = {text width=6cm, align=center},
    legend style={at={(0.375,0.75)}, anchor=north west, legend cell align=left}]
    \addplot[fill=yellow!40, error bars/.cd, y dir=both, y explicit]
        plot coordinates {
            (BLIP, 893) 
            (I0T$_{\text{async}}$, 354) 
            ($\text{I0T}_{\text{post}}$, 353)};
    \legend {$\#$ Parameters $\downarrow$};
\end{axis}
\begin{axis}[
    width  = 0.85\columnwidth,
    height = 5cm,   
    ybar,
    bar width = 3.0mm,
    bar shift = 0.0cm,
    enlarge x limits = 0.20,
    ylabel style = {align=center},
    ymin = 0,
    ymax = 1,
    ytick pos = right,
    ytick align = outside,
    symbolic x coords = {BLIP,I0T$_{\text{async}}$,$\text{I0T}_{\text{post}}$},
    axis y line*=right,
    xtick = data,
    xtick style={draw=none},
    xtick pos = bottom,
    xtick align= outside,
    xticklabel style = {rotate=0, text width=2.0cm, align=center},
    xlabel style = {text width=6cm, align=center},
    legend style={at={(0.24,0.95)}, anchor=north west, legend cell align=left}]
    \addplot[fill=red!40, error bars/.cd, y dir=both, y explicit]
        plot coordinates {
            (BLIP, 0.9570)
            (I0T$_{\text{async}}$, 0.4795)
            ($\text{I0T}_{\text{post}}$, 0.0102)};
    \legend {Centroid Distance $\downarrow$};
\end{axis}
\begin{axis}[
    width  = 0.85\columnwidth,
    height = 5cm,   
    ybar,
    bar width = 3.0mm,
    bar shift = 0.4cm,
    enlarge x limits = 0.20,
    ylabel style = {align=center},
    ymin = 0,
    ymax = 1,
    ytick = {0.0, 0.2, 0.4, 0.6, 0.8, 1.0},
    ytick pos = right,
    ytick align = outside,
    symbolic x coords = {BLIP,I0T$_{\text{async}}$,$\text{I0T}_{\text{post}}$},
    xtick=\empty,
    xticklabel=\empty,
    axis y line*=right,
    legend style={at={(0.175,1.15)}, anchor=north west, legend cell align=left}]
    \addplot[fill=blue!40, error bars/.cd, y dir=both, y explicit]
        plot coordinates {
            (BLIP, 0.5020)
            (I0T$_{\text{async}}$, 0.5333) 
            ($\text{I0T}_{\text{post}}$, 0.5397)};
    \legend {Relative Correlation $\uparrow$};
\end{axis}
\end{tikzpicture}
\caption{Comparison of non-CLIP-based model BLIP and ours on the efficiency and performances.}
\label{fig:blip_comparison}
\end{figure}

\subsection{Comparison across Different Methods}

As can be observed in Table~\ref{tab:main_comparison}, $\text{I0T}_{\text{post}}$ and I0T$_{\text{async}}$ can substantially reduce the modality gap without hurting the overall downstream performances, maintaining the first and second rankings\footnote{We report the mean ranking of each model measured across six downstream tasks and two modality gap metrics to provide a holistic perspective.}. While CLIP shows a severe modality gap with high CD and LS scores (0.7642 and 0.9985, respectively), indicating significant separation in the latent space between image and text embeddings, $\text{I0T}_{\text{post}}$ reduces the modality gap to almost zero with significantly low CD and LS scores of 0.0102 and 0.5374. This is notably better across tasks compared to a competitive post-hoc method, MG$_{\lambda=0.375}$ \cite{liang2022mind}, which also achieves a low severity level of modality gap. Note that while the MG post-hoc method requires a tuning of $\lambda$ ($\lambda=-0.5,0.375,0.5$) to achieve the low modality gap, our post-hoc method does not require hyperparameter tuning. Meanwhile, I0T$_{\text{async}}$ reduces this gap to a moderate level with CD and LS scores of 0.4795 and 0.9960.

We also ensure that the I0Ts do not significantly hurt semantic representations when achieving the goal of reducing the modality gap. Table~\ref{tab:main_comparison} indicates our methods especially achieve competitive retrieval scores in Flickr30k \cite{plummer2015flickr30k}; I0T$_{\text{async}}$: 72.50\% and 73.80\% for I2T and T2I retrieval scores and $\text{I0T}_{\text{post}}$: 73.30\% and 76.30\%. We emphasize that $\text{I0T}_{\text{post}}$ shows very close performances on the Flickr-Expert/CF dataset \cite{hodosh2013framing} compared to PAC-S \cite{sarto2023positive}, the state-of-art contrastive-based evaluation metric, and scores 4.46\%  higher in CIFAR classification \cite{krizhevsky2009learning}. Furthermore, I0T$_{\text{async}}$ shows enhanced downstream performances than CLOOB \cite{furst2022cloob} and Unif-Align \cite{wang2020understanding} while achieving a similar moderate level of modality gap. Figure~\ref{fig:blip_comparison} illustrates that I0Ts also achieve 94.68\% and 47.74\% lower CD compared to BLIP with 2.5 times fewer parameters while achieving similar correlation performances on Flickr-Expert. The results of varying batch sizes and ResNet-based CLIPs are in Appendix~\ref{app:add_exp}.

\begin{figure}[t!]
    \centering
    \includegraphics[width=\columnwidth]{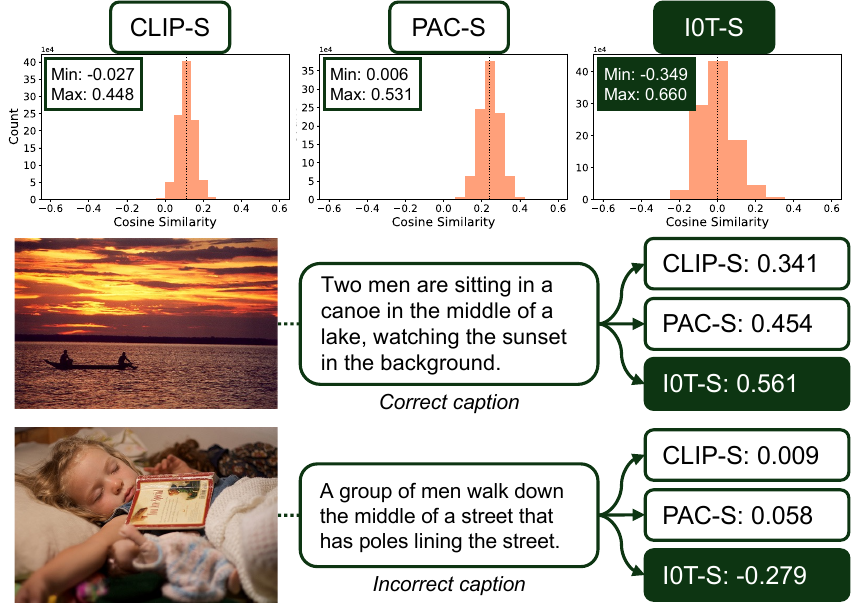}
    \caption{A wider range of cosine similarity distribution with mean close to 0 using I0T-S compared to CLIP-S and PAC-S without the scaling factor (\textit{i.e.}, $\omega=1$), contributing to more explainable similarity scores for positive and negative pair of image and caption.}
    \label{fig:similarity_score}
\end{figure}

\subsection{Applicability as an Automatic Reference-free Evaluation Metric}
In Figure~\ref{fig:similarity_score}, our I0T-S, built upon I0T$_\text{post}$, shows a non-skewed, wider cosine similarity distribution from -0.349 to 0.660, in comparison to the popular automatic reference-free image captioning evaluation metric, CLIPScore (CLIP-S) \cite{hessel2021clipscore} and PAC-S \cite{sarto2023positive}. This indicates that I0T-S is a more intuitive, explainable metric than these conventional methods, assigning a higher similarity score for the positive pair with the correct caption (+ 21.0\% than CLIP-S) and a lower similarity score for the negative pair with the incorrect caption (-20.0\% than CLIP-S).

This illustrates the necessity and benefits of reducing the modality gap, especially when we use CLIP as a reference-free evaluation metric. While CLIP-S and PAC-S typically use the scaling factor ($\omega$) 2.5 on the raw cosine similarity scores to improve numerical readability \cite{sarto2023positive}, this scaling merely enlarges the cosine similarity distribution without altering the image and text embeddings in the latent space. For instance, the cosine similarity distribution of PAC-S \cite{sarto2023positive} has a minimum \textit{positive} value of 0.006 for an incorrect image and caption pair. Scaling this value with $\omega =$ 2.5 yields 0.015, which unintuitively assigns a higher similarity score to an incorrect pair by scaling. In contrast, I0T-S does not require scaling due to the reduced modality gap property; thus, I0T-S not only shows a high relative correlation with human ranking but also yields interpretable absolute values of similarity scores.

\section{Discussion}

\subsection{The Relationship between Modality Gap and Downstream Performances}

$\text{I0T}_{\text{post}}$ with the lowest modality gap severity level achieves the highest task performance in image-text retrieval (Table~\ref{tab:main_comparison}). However, at the same time, it does not always score the highest performances for classification and correlation tasks. We emphasize that there is no direct causal relationship between the modality gap and downstream performances, similar to ongoing discussions in recent works \cite{liang2022mind, jiang2023understanding, ramasinghe2024accept, schrodi2024two}. Specifically, \citealp{liang2022mind} states that sometimes a “larger gap” can help improve zero-shot learning performances, \citealp{jiang2023understanding} empirically shows unguaranteed downstream performances when reducing modality gap. Our work does not claim any relationships between modality gap reduction and performance improvement in line with these studies. Rather, we show that our methods could significantly reduce the modality gap without hurting overall downstream performances.

We believe the best of both perspectives can be achieved through the presence of our plug-in-play module of the first stage, which solely focuses on enhancing semantic features in the separation of stage 2, adding single batch normalization layers for each encoder. Although $\text{I0T}_{\text{async}}$ does not mitigate the modality gap to near zero due to distribution differences between training and test samples, it still reduces to the moderate level. This suggests we could use a pre-computed embedding average from a subset of the training dataset as another solution when dealing with the modality gap if we also do not have enough resources for training. However, we emphasize that unlike recently introduced learning methods \cite{wang2020understanding, eslami2024mitigate, xia2024achieving}, our learnable approach does not change (but mostly improve) much of the original embeddings with trainable BN layers separately added to pre-trained encoders.

\subsection{Why is Batch Normalization Effective in Reducing the Modality Gap?}

Here, we find that peak activations across a few dimensions for each modality encoder are the main reasons for the large modality gap. However, we also observe that simply clipping these peak activations does not help to reduce the gap. Thus, instead of directly linking the \textit{modality-specific characteristics} into only peak activations, we aim to remove the aggregated mean/std statistics of normalized embedding activations for each modality. This is possible due to the consistently similar values of means and minuscule standard deviations over data samples per modality and minuscule standard deviations across all hidden dimensions, which can be effectively learned using separate BN layers. In addition, other normalizations, such as LN, do not help reduce the modality gap effectively since the objective of LN is not linked to the modality gap.

Furthermore, our asynchronous strategies of applying post-hoc and training BN methods on frozen encoders allow the model to significantly reduce the modality gap while preserving semantic representations of embeddings. Thus, although BN layers were conventionally thought of as one of the normalization strategies in the past, we could reinterpret these as effective strategies for mitigating the modality gap. 

\section{Conclusion}

In this study, we present the I0T framework that can effectively reduce the modality gap between image and text embeddings while preserving the semantic representations. We first introduce a simple post-hoc embedding standardization method of reducing the gap to the close-zero value ($\text{I0T}_{\text{post}}$) and also provide a novel training strategy using separate batch normalization layers for each modality (I0T$_{\text{async}}$). I0Ts show effectiveness in both modality gap and downstream performances compared to the other seven CLIP-based models and BLIP with no additional and 10M extra training parameters for $\text{I0T}_{\text{post}}$ and I0T$_{\text{async}}$, respectively. We believe this work will guide and inspire future research to address the modality gap further, an area less explored than improving downstream performances.

\section{Limitations}

While I0Ts demonstrate significant improvements in reducing the modality gap, $\text{I0T}_{\text{post}}$ relies on the entire test dataset, which may not be practical for single-sample inference when sufficient data is unavailable. To address this limitation, we introduce $\text{I0T}_{\text{async}}$, which reduces the modality gap without requiring access to the test dataset. However, the modality gap achieved with $\text{I0T}_{\text{async}}$ does not entirely reach the near-zero level, similar to existing methods such as CLOOB and Unif-Align. Here, we provide a simple baseline method using the existing BN layers. We leave it as a future study to explore different learning methods and additional modalities (\textit{e.g.}, audio and video) for reducing the modality gap.

\section{Ethical Statement}
Misusing our proposed metric, I0T-S, as the reference-free evaluation metric could bring a potential risk. However, we believe this applies to every reference-free metric since I0T-S is built upon the widely used ClIP-S.



\newpage
\appendix
\onecolumn

\section{Categorization of Modality Gap Severity Levels}\label{app:categ}
We visualize the normalized image (red) and text (blue) embeddings on the 3D sphere using UMAP \cite{mcinnes2018umap} setting the output metric as haversine\footnote{We fix the seed as 42 for the reproducibility purpose.}. The dots on the scatter plot represent shifted CLIP embeddings randomly sampled with replacement from 1k validation set of Flickr30k \cite{plummer2015flickr30k}. Note that there are 100 dots for each $\lambda$ (color) ranging from 0.4 to 1.0 \cite{liang2022mind}. We interpolate these dots, fitting piecewise linear functions with the Scipy (optimize) package for (1) linear separability ($y_1$) \textit{vs}. centroid distance ($x$), and (2) minimum cosine distance ($y_2$) \textit{vs}. centroid distance ($x$). The resulting functions are as follows: 

$$
y_1 =
\begin{cases}
4.53x + 0.97-(4.53)(0.19), & \text{if }x < 0.19 \\
0.04x + 0.97-(0.04)(0.19), & \text{if }x \geq 0.19
\end{cases}
$$

$$
y_2 =
\begin{cases}
0.17x + 0.39-(0.17)(0.63), & \text{if }x < 0.63 \\
0.75x + 0.39-(0.75)(0.63), & \text{if }x \geq 0.63
\end{cases}
$$

\section{Analyzing Normalized Embedding Activations}\label{app:norm_act}
\begin{figure}[h!]
    \centering
    \includegraphics[width=0.5\columnwidth]{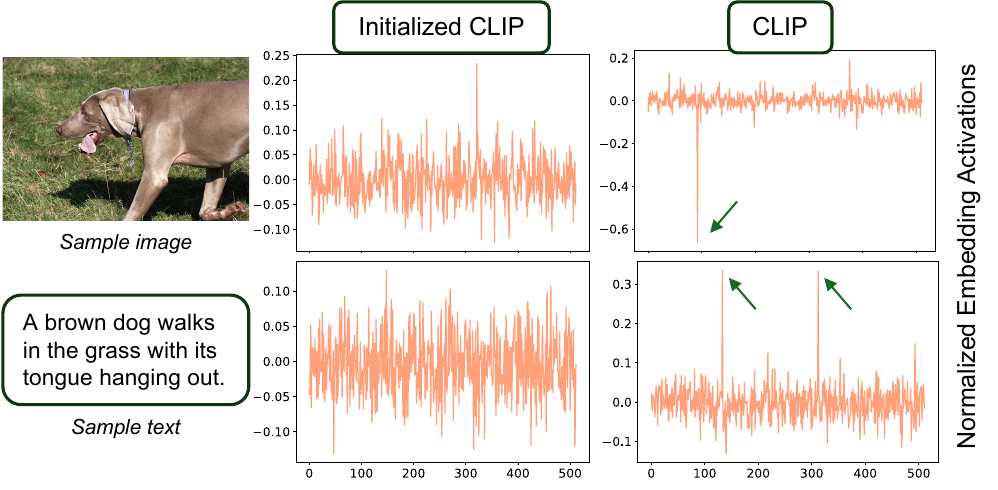}
    \caption{The existence of positive and negative peak activations in normalized embedding activations (pointed by arrows) for a sample image and corresponding caption.}
    \label{fig:outlier_activations}
\end{figure}

\section{Proof of Claim}\label{app:proof}
We derive an upper bound of the absolute value of the cosine similarity of normalized image embedding activations ($\textbf{x}=[x_1, x_2, ..., x_d]$) and normalized text embedding activations ($\textbf{y}=[y_1, y_2, ..., y_d]$), $|\text{cos}(\textbf{x}, \textbf{y})|$, where each activation contains one and two peak activations, $p$ and $q$'s at different dimensions. Since we assume that the other non-peak activations are uniformly distributed, $x_i = x_j$ and $y_i = y_j$ for $i, j \in \{1, 2, ..., d\}$ such that $i \neq j$. Then,

\begin{gather*}
|\text{cos}(\textbf{x}, \textbf{y})| \leq \sum_{i=1}^{d} |x_i y_i| \\
= |x_{t_1}q| + |x_{t_2}q| + |py_{t_3}| + \sum_{i \notin \{t_1, t_2, t_3\}}|{x_i y_i}| \\
= 2|q|\sqrt{\frac{1-p^2}{d-1}} + |p|\sqrt{\frac{1-2q^2}{d-2}} \\ + (d-3)\sqrt{(\frac{1-p^2}{d-1})(\frac{1-2q^2}{d-2})} \\
(\because \sum_{i=1}^{d} x_i^2 = \sum_{i=1}^{d} y_i^2 = 1)
\end{gather*}

Thus, $\text{lim}_{d \rightarrow \infty} |\text{cos}(\textbf{x}, \textbf{y})| \leq \sqrt{(1-p^2)(1-2q^2)}$ and $|\text{cos}(\textbf{x}, \textbf{y})| \rightarrow 0$ as $d \rightarrow \infty$, $p \rightarrow 1$, and $q \rightarrow 1/\sqrt{2}$.

\begin{table}[t!]
\centering
\resizebox{0.6\columnwidth}{!}{%
\begin{tabular}{c|cc|cc|cc|cc}
\toprule
\textit{Training} & \multicolumn{2}{c|}{LAION} & \multicolumn{2}{c|}{COCO} & \multicolumn{2}{c|}{SAM} & \multicolumn{2}{c}{\textit{All}} \\
\textit{Datasets} & \multicolumn{2}{c|}{(558k)} & \multicolumn{2}{c|}{(118k)} & \multicolumn{2}{c|}{(570k)} & \multicolumn{2}{c}{(1.2m)} \\
\midrule

\textit{Models} & LC & LCO & LC & LCO & LC & LCO & LC & LCO \\
\midrule

\multicolumn{1}{c|}{I2T} & 65.30 & 65.70 & \underline{71.90} & \textbf{72.50} & 69.70 & 71.7 & 69.4 & 71.0\\
\multicolumn{1}{c|}{T2I} & 73.30 & 73.50 & \underline{75.00} & \textbf{75.90} & 72.00 & 74.7 & 73.5 & 73.7\\
\midrule
 
\multicolumn{1}{c|}{CIFAR} & \textbf{67.21} & 66.49 & 65.03 & 64.46 & 65.26 & 65.23 &66.67& \underline{64.73}\\
\multicolumn{1}{c|}{Bird} & 5.46 & \textbf{6.06} & 5.43 & \underline{5.81} & 5.50 & 5.64 &5.46& 5.53\\
\midrule

\multicolumn{1}{c|}{Expert} & \textbf{53.78} & \underline{53.59} & 53.21 & 51.42 & 52.25 & 53.39 &53.24& 53.03\\
\multicolumn{1}{c|}{CF} & 34.75 & 35.16 & 34.86 & \textbf{35.17} & 33.71 & 34.92 &34.27& \underline{35.11}\\
\bottomrule
\end{tabular}}
\caption{Downstream performances of Long-CLIP (LC) and Long-CLIP-only (LCO) fine-tuned on three independent subsets of ShareGPT4V (LAION, COCO, SAM) and all together. The \textbf{bolded} and \underline{underlined} values indicate the best and the second-best performances.}
\label{tab:dataset_ablation_comparison}
\end{table}

\begin{table}[t!]
\centering
\resizebox{0.35\columnwidth}{!}{%
\begin{tabular}{c|cc|cc}
\toprule

\multicolumn{1}{c|}{} & I2T & T2I & I2T & T2I \\

\multicolumn{1}{c|}{Models} & \multicolumn{2}{c|}{5 captions}  &  \multicolumn{2}{c}{1 caption}  \\ 

\midrule

CLIP & 78.5 & 58.7 & 69.6 & 67.1  \\
$\text{I0T}_{\text{async}}$  & \textbf{80.6} & \underline{63.3} & \underline{72.5}  & \underline{73.8} \\ 
$\text{I0T}_{\text{post}}$ & \underline{80.0} & \textbf{65.6} & \textbf{73.3} & \textbf{76.3}  \\
\bottomrule
\end{tabular}}
\caption{Comparison of retrieval scores using conventional five captions and proposed one caption. The \textbf{bolded} and \underline{underlined} values indicate the best and the second-best performances.}
\label{tab:five_captions}
\end{table}

\begin{table}[t!]
\centering
\begin{tcolorbox}[colback=white, colframe=black, boxsep=3pt, left=3pt, right=3pt, top=3pt, bottom=3pt, width=0.4\columnwidth, sharp corners]
a photo of a [class] \\
a blurry photo of a [class] \\
a black and white photo of a [class] \\
a low contrast photo of a [class] \\
a high contrast photo of a [class] \\
a bad photo of a [class] \\
a good photo of a [class] \\
a photo of a small [class] \\
a photo of a big [class] \\
a photo of the [class] \\
a blurry photo of the [class] \\
a black and white photo of the [class] \\
a low contrast photo of the [class] \\
a high contrast photo of the [class] \\
a bad photo of the [class] \\
a good photo of the [class] \\
a photo of the small [class] \\
a photo of the big [class]
\end{tcolorbox}
\caption{18 templates for classifying images in CIFAR100.}
\label{tab:classification_prompts}
\end{table}

\begin{figure}[t!]
    \centering
    \includegraphics[width=0.5\columnwidth]{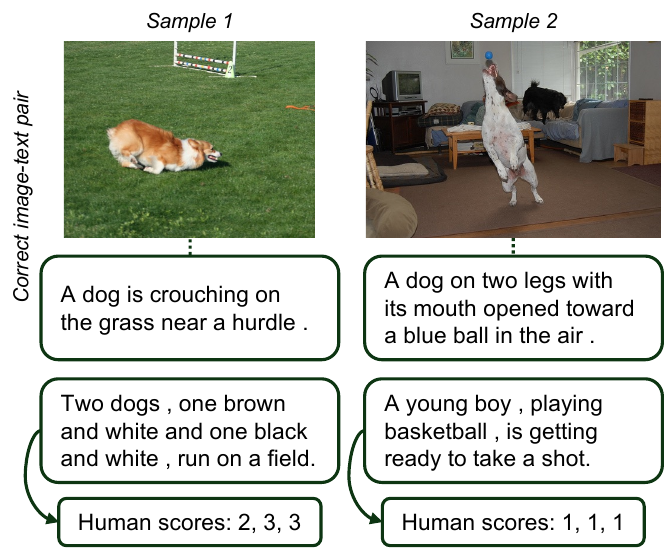}
    \caption{Sample image-caption pairs and corresponding human scores from Flickr8k-Expert dataset.}
\label{fig:hum_scores_example}
\end{figure}

\section{Experimental Design Details}\label{app:exp_deta}

\subsection{Fine-tuning Dataset Selection}

To investigate the effect of different fine-tuning datasets on downstream performances, we assess Long-CLIP (LC) \cite{zhang2024long} and Long-CLIP-only (LCO) using LAION, COCO, SAM, and a combined set, as illustrated in Table \ref{tab:dataset_ablation_comparison}. To guarantee statistical robustness and reproducability purposes, each dataset is fine-tuned using three random seeds (7, 42, and 71). We generally find that LCO performs better than LC, suggesting that the alignment of short captions and the re-constructed images does not help improve the models' ability of enhanced semantic understanding.

Also, we find that fine-tuning these models on COCO shows the best performances on I2T and T2I retrieval tasks, achieving scores of 72.5\% and 75.9\%, respectively, outperforming models fine-tuned on LAION, SAM, and the combined dataset with the lowest number of training samples. This success is possibly attributed to COCO's diverse and detailed image-caption pairs \cite{lin2014microsoft}, which closely match the characteristics of evaluation datasets. Although the number of samples in the combined dataset is approximately ten times larger, the performances of models fine-tuned on the combined dataset are not improved proportionally to the number of datasets. Hence, we select COCO as the final fine-tuning dataset to ensure strong performances across various downstream applications with high efficiency during training.

\subsection{Evaluation Downstream Tasks}

To perform the image-text retrieval task, we use the Kaparthy validation split \cite{karpathy2015deep} of \textbf{Flickr30k} \cite{plummer2015flickr30k} (1k images), following the conventional short-caption retrieval task as in \citealp{zhang2024long} and not long-caption retrieval task since our comparison models that are built upon CLIP (\textit{e.g.}, CLOOB and PAC-S) can handle up to 77 tokens. We use the first caption for each image to calculate T2I and I2T R@1 scores, different from previous studies \cite{li2022blip, goel2022cyclip} that use all five captions, resulting in an imbalance between I2T and T2I scores (\textit{see} Table~\ref{tab:five_captions}). Also, there exist consistent relative trends across models between the two scoring systems. The Flickr30k dataset is also used to evaluate the modality gap.

We evaluate the zero-shot image classification ability of models using \textbf{CIFAR100} \cite{krizhevsky2009learning} (10k images, 100 classes), and \textbf{Birdsnap} \cite{berg2014birdsnap} (1,857 images, 500 classes) with 18 templates and one template, respectively, and report class-weighted balanced accuracy. Below, we list the exact templates we used for classifying 100 classes in the CIFAR100 \cite{krizhevsky2009learning} dataset (Table~\ref{tab:classification_prompts}). We average the text embeddings for each class (name) over the templates to calculate the similarity between image and classes for zero-shot classification. For the Birdsnap \cite{berg2014birdsnap}, we use one template: ``a photo of a [class], a type of bird" \cite{furst2022cloob}.

Lastly, we assess the correlation ability of models using \textbf{Flickr8k-Expert} (1k images) and \textbf{Flickr8k-CF} (1k images) \cite{hodosh2013framing}, which contain sentences with human judgment scores ranging from 1 to 5. In Figure~\ref{fig:hum_scores_example}, we show two samples of human scores, each from Flickr8k-Expert and Flickr8k-CF \cite{hodosh2013framing} datasets. We report Kendall's correlation coefficient ($\tau_{b}$) between average human scores and cosine similarity between image and text embeddings.

\subsection{Comparison Methods}

In this study, we compare five state-of-the-art methods with our proposed method: \textbf{Mind-the-Gap} (MG), \textbf{CLOOB}, \textbf{Unif-Align}, \textbf{PAC-S}, and \textbf{BLIP}. \textbf{Mind-the-Gap} (MG)
\cite{liang2022mind}  mitigates the modality gap by adjusting image and text embeddings post-training through the subtraction and addition of $\lambda \triangle_{\text{CD}}$. \textbf{CLOOB} \cite{furst2022cloob} utilizes modern Hopfield networks \cite{ramsauer2021hopfield} to retrieve embeddings that store covariance structures and InfoLOOB objective \cite{poole2019variational}. \textbf{Unif-Align} \cite{wang2020understanding}, as tested in \citealp{fahim2024its}, enhances the uniformity within and the alignment between image and text embeddings. For CLOOB and Unif-Align, we reproduce results using our baseline Long-CLIP-only fine-tuning with COCO. \textbf{PAC-S} \cite{sarto2023positive} is a CLIP-based model proven effective in correlating human judgments on images using positive-augmented contrastive learning loss with synthetically generated images and the correspondingly generated texts. \textbf{BLIP} \cite{li2022blip}, a larger non-CLIP-based vision-language model, is known for its strong performance across various multimodal tasks. We evaluate PAC-S and BLIP using the provided checkpoints.

\section{\texorpdfstring{I0T$_{\text{async}}$: Learnable Method to Reduce Modality Gap}{Learnable Method to Reduce Modality Gap in Building}}\label{app:alg}

\begin{center}
\begin{minipage}{0.6\textwidth}
\begin{algorithm}[H]
\caption{Extraction of Image/Text Embedding Representations in PyTorch-like Style}
\label{alg:cap}
\begin{algorithmic}[1]
\State \textbf{Input} $I$: image \& $T$: text-based caption
\State \textbf{Require} $\mathcal{E}_{\text{img}}$/$\mathcal{E}_{\text{txt}}$: vision/text encoder, $\text{BN}_{\text{img}}$/$\text{BN}_{\text{txt}}$: $\text{BN}$ layer for images/texts, and AddBatchNorm: boolean
\State \textbf{Output} $\mathcal{E}^{I}$/$\mathcal{E}^{T}$: image/text embeddings
\State
\Function {EncodeImage}{$I$}
\State $\mathcal{E}^{I} = \mathcal{E}_{\text{img}}(I)$
\If{AddBatchNorm} 
    \State $\textbf{x} = \mathcal{E}^{I} / \text{Norm}(\mathcal{E}^{I}, \text{dim}=1)$
    \State $\mathcal{E}^{I} = \text{BN}_{\text{img}}^{I}(\textbf{x})$
\EndIf 
\State \Return $\mathcal{E}^{I}$
\EndFunction 
\State
\Function {EncodeText}{$T$}
\State $\mathcal{E}^{T} = \mathcal{E}_{\text{txt}}(T)$
\If{AddBatchNorm}
    \State $\textbf{y} = \mathcal{E}^{T} / \text{Norm}(\mathcal{E}^{T}, \text{dim}=1)$
    \State $\mathcal{E}^{T} = \text{BN}_{\text{txt}}^{T}(\textbf{y})$
\EndIf 
\State \Return $\mathcal{E}^{T}$
\EndFunction 
\end{algorithmic}
\end{algorithm}
\end{minipage}
\end{center}

\section{Additional Experiments}\label{app:add_exp}

\paragraph{Ablation on MCSIE}
Table~\ref{tab:mcsie} demonstrates that training I0T$_{\text{async}}$ without MCSIE results in moderate modality gap severity but performs slightly lower in retrieval tasks than I0T$_{\text{async}}$ with MCSIE, highlighting the benefit of using MCSIE when training independent BN layers.

\begin {figure}[t!]
\centering
\begin{tikzpicture}
\begin{axis}[
    width  = 0.5\columnwidth,
    height = 5cm,   
    ybar,
    bar width = 3.0mm,
    enlarge x limits=0.18,
    ylabel style = {align=center},
    ymin = 0.4,
    ymax = 1.0,
    ytick = {0.4, 0.5, 0.6, 0.7, 0.8, 0.9, 1.0},
    ytick pos = left,
    ytick align = outside,
    symbolic x coords = {\text{32$\times$2}, \text{64$\times$2}, \text{128$\times$2}, \text{256$\times$2}},
    xtick = data,
    xlabel = Batch Sizes,
    nodes near coords, 
    every node near coord/.append style={rotate=90,
    anchor=west,/pgf/number format/.cd,/pgf/number format/fixed,/pgf/number format/fixed zerofill,/pgf/number format/precision=4},
    xtick style={draw=none},
    xtick pos = bottom,
    xtick align= outside,
    xticklabel style = {rotate=0, text width=2.0cm, align=center},
    xlabel style = {text width=6cm, align=center},
    legend columns=-1,
    legend style={at={(-0.01,1.15)}, anchor=north west, legend cell align=left,/tikz/every even column/.append style={column sep=0.34cm}}]
    \addplot[fill=red!40, error bars/.cd, y dir=both, y explicit]
        plot coordinates {
            (\text{32$\times$2}, 0.4776)
            (\text{64$\times$2}, 0.4795)
            (\text{128$\times$2}, 0.4856)
            (\text{256$\times$2}, 0.4832)};
    \addplot[fill=cyan!40, error bars/.cd, y dir=both, y explicit]
        plot coordinates {
            (\text{32$\times$2}, 0.7250)
            (\text{64$\times$2}, 0.7250)
            (\text{128$\times$2}, 0.7250)
            (\text{256$\times$2}, 0.7280)};
    \addplot[fill=blue!40, error bars/.cd, y dir=both, y explicit]
        plot coordinates {
            (\text{32$\times$2}, 0.5390)
            (\text{64$\times$2}, 0.5333)
            (\text{128$\times$2}, 0.5344)
            (\text{256$\times$2}, 0.5347)};
    \legend {Centroid Dist $\downarrow$, I2T $\uparrow$, Corr $\uparrow$};
\end{axis}
\end{tikzpicture}
\caption{Effect of varying batch sizes on centroid distance, retrieval, and correlation performances of I0T$_{\text{async}}$.}
\label{fig:batchsize_comparison}
\end{figure}
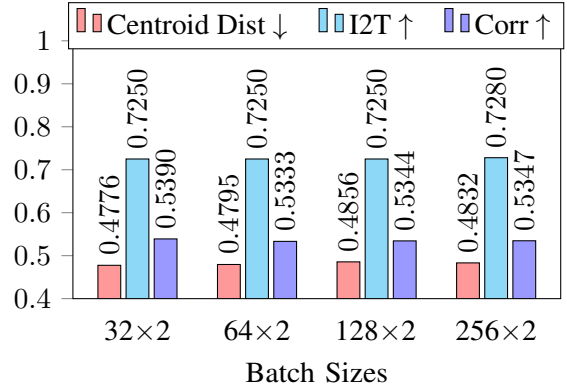

\paragraph{Effectiveness of Varying Batch Sizes}
We examine the effect of different batch sizes on key metrics such as centroid distance, I2T on Flickr30K, and human judgment correlation on Flickr30k-Expert using I0T$_{\text{async}}$ (Figure \ref{fig:batchsize_comparison}). Specifically, we explore batch sizes of 32, 64, 128, and 256, each scaled by the number of GPU devices utilized. We find that variations in batch sizes have a minimal impact on training outcomes, captured with no significant change in downstream performances and modality gap scores. This affirms that the batch size is not a critical factor when training I0T$_{\text{async}}$.

\paragraph{ResNet-based CLIP}
In Table~\ref{tab:resnet_comparison}, we show that our post-hoc method is also applicable to CLIP (RN50) \cite{radford2021learning}. Clearly, CLIP$_\text{post}$ shows the lowest modality score and better retrieval scores than Mind-the-Gap (MG) baselines \cite{liang2022mind}. Whereas the T2I score is 3.1 higher than that of CLIP, the I2T score is 3.3 lower than the CLIP I2T score.

\paragraph{Computational Cost}
I0T$_{\text{post}}$ is a training-free method that requires no additional training cost. I0T$_{\text{async}}$ requires only 10M additional model parameters to the original CLIP and requires 6.7 GiB of GPU memory without considering the model load in our experimental setting. All the training and evaluation experiments are conducted using two NVIDIA RTX A4000s.

\begin{table}[t!]
\centering
\resizebox{0.45\columnwidth}{!}{%
\begin{tabular}{c|cc|cc}
\toprule

\multicolumn{1}{c|}{} & Centroid & Linear & \multicolumn{2}{c}{Retrieval} \\

\multicolumn{1}{c|}{Models} & Dist. &  Sep. & I2T & T2I  \\ 

\midrule

BN* wo/ MCSIE & 0.442 &	0.999 &	70.9 &	72.3  \\
BN* w/ MCSIE & 0.479 &	0.996 & 72.5 &	73.8 \\ 
\bottomrule
\end{tabular}}
\caption{Comparison of BN* without and with proposed MCSIE approach.}
\label{tab:mcsie}
\end{table}

\begin{table}[t!]
\centering
\resizebox{0.5\columnwidth}{!}{%
\begin{tabular}{c|ccc|cc}
\toprule
\multicolumn{1}{c|}{} & \multicolumn{3}{c|}{Modality Gap} & \multicolumn{2}{c}{Performances} \\
 
\multicolumn{1}{c|}{Models} & Centroid & Linear & Sev. & \multicolumn{2}{c}{Retrieval $\uparrow$} \\

\multicolumn{1}{c|}{} & Dist. $\downarrow$ & Sep. $\downarrow$ & Level $\downarrow$ & \multicolumn{1}{c}{I2T} & \multicolumn{1}{c}{T2I} \\ 
\midrule
 
CLIP  & 0.7647 & 0.9964 & \cellcolor[HTML]{EC7063} \textit{sev} & \textbf{71.30} & \underline{67.20} \\ 
$\text{MG}_{\lambda=0.375}$ & \underline{0.0214} & \underline{-1.4094} & \cellcolor[HTML]{FDEDEC} \textit{low} & 47.00 & 51.20 \\
$\text{MG}_{\lambda=0.5}$  & 0.2481 & 0.9631 & \cellcolor[HTML]{F5B7B1} \textit{mod} & 37.10 & 42.80 \\
$\text{MG}_{\lambda=-0.5}$  & 1.3799 & 0.9996 & \cellcolor[HTML]{EC7063} \textit{sev} & 42.60 & 50.90  \\ 
\midrule
CLIP$_{\text{post}}$  & \textbf{0.0097} & \textbf{-1.8497} & \cellcolor[HTML]{FDEDEC} \textit{low} & \underline{68.00} & \textbf{70.30} \\ 
\bottomrule
\end{tabular}}
\caption{Comparison of modality gap and downstream performances across different ResNet-based CLIP models ($\#$ Param: 255m). The \textbf{bolded} and \underline{underlined} values indicate the best and the second-best performances.}
\label{tab:resnet_comparison}
\end{table}

\end{document}